\documentclass[journal]{IEEEtran}
\usepackage{cite}
\usepackage{mathtools}
\usepackage{xfrac}
\usepackage{color}
\usepackage{stfloats}
\usepackage[noend]{algpseudocode}
\usepackage{algorithmicx,algorithm}
\usepackage{subfigure}
\newcommand{\red}[1]{\textcolor{red}{#1}}
\newcommand{\green}[1]{\textcolor{green}{#1}}
\newcommand{\blue}[1]{\textcolor{blue}{#1}}
\ifCLASSINFOpdf
\graphicspath{{../pdf/}}
\DeclareGraphicsExtensions{.pdf,.jpeg,.png}
\else
\graphicspath{{../eps/}}
\DeclareGraphicsExtensions{.eps}
\fi

\begin{document}
\title{SurroundNet: Towards Effective Low-Light Image Enhancement}

\author{\IEEEauthorblockN{Fei Zhou,
Xin Sun, ~\IEEEmembership{Member,~IEEE,}
Junyu Dong, ~\IEEEmembership{Member,~IEEE,}
Haoran Zhao,
Xiao Xiang Zhu, ~\IEEEmembership{Fellow,~IEEE}
\thanks{This work was supported in part by National Natural Science Foundation of China under Project Nos. 61971388 and U1706218, and Key Research and Development Program of Shandong Province (No. GG201703140154).}
	\thanks{}
}

}

\markboth{SUBMITTED TO }%
{Shell \MakeLowercase{\textit{et al.}}: Bare Demo of IEEEtran.cls for IEEE Transactions on Magnetics Journals}

\IEEEtitleabstractindextext{%
\begin{abstract}
Although Convolution Neural Networks (CNNs) has made substantial progress in the low-light image enhancement task, one critical problem of CNNs is the paradox of model complexity and performance. This paper presents a novel SurroundNet which only involves less than 150$K$ parameters (about 80-98 percent size reduction compared to SOTAs) and achieves very competitive performance. The proposed network comprises several Adaptive Retinex Blocks (ARBlock), which can be viewed as a novel extension of Single Scale Retinex in feature space. The core of our ARBlock is an efficient illumination estimation function called Adaptive Surround Function (ASF). It can be regarded as a general form of surround functions and be implemented by convolution layers. In addition, we also introduce a Low-Exposure Denoiser (LED) to smooth the low-light image before the enhancement. We evaluate the proposed method on the real-world low-light dataset. Experimental results demonstrate that the superiority of our submitted SurroundNet in both performance and network parameters against State-of-the-Art low-light image enhancement methods. Code is available at https: github.com/ouc-ocean-group/SurroundNet.
\end{abstract}

}

\maketitle

\IEEEdisplaynontitleabstractindextext

\IEEEpeerreviewmaketitle

\section{introduction}

\IEEEPARstart{S}{hadow} or a pedestrian? It makes no difference when driving at night until your headlights swept it. Poor illumination condition can appear anywhere and bring complex image degradation, such as signal-dependent noise, low contrast and all that. It hurts the performance of any vision system, both human and computer. To handle such problem, various theories were proposed which can be divided into two categories in general: the histogram-based and Retinex-based methods \cite{lv2018mbllen}. In recent years, Convolutional Neural Networks (CNNs) produce compelling results in low-level image processing such as Denoiser\cite{mao2016image} and Super-Resolution\cite{2014Learning}. It makes CNN-based methods become a good alternative and popular route for image enhancement. Due to the data-driven learning and deep structure, the CNN-based methods can achieve better performance in some specific datasets compared with conventional ones. And some works have successfully combined the Retinex and CNNs \cite{wei2018deep, shen2017msr, wang2019progressive}. Despite the success, all of those Retinex-CNN methods drop the surround function from conventional Retinex methods like Single Scale Retinex (SSR) \cite{jobson1997properties} and stack convolution layers to obtain the illumination map. It is reasonable to some extent that the data-driven illumination estimation can restrain the halo artifacts and has better robustness. However, different from surround function, there is hardly prior knowledge about illumination in CNNs. To achieve the same ends, networks have to utilize numerous redundant parameters to learn the illumination map. Therefore, it inevitably increases model complexity and slows the inference speed.

\begin{figure}[t]
\centering
\includegraphics[scale=0.45]{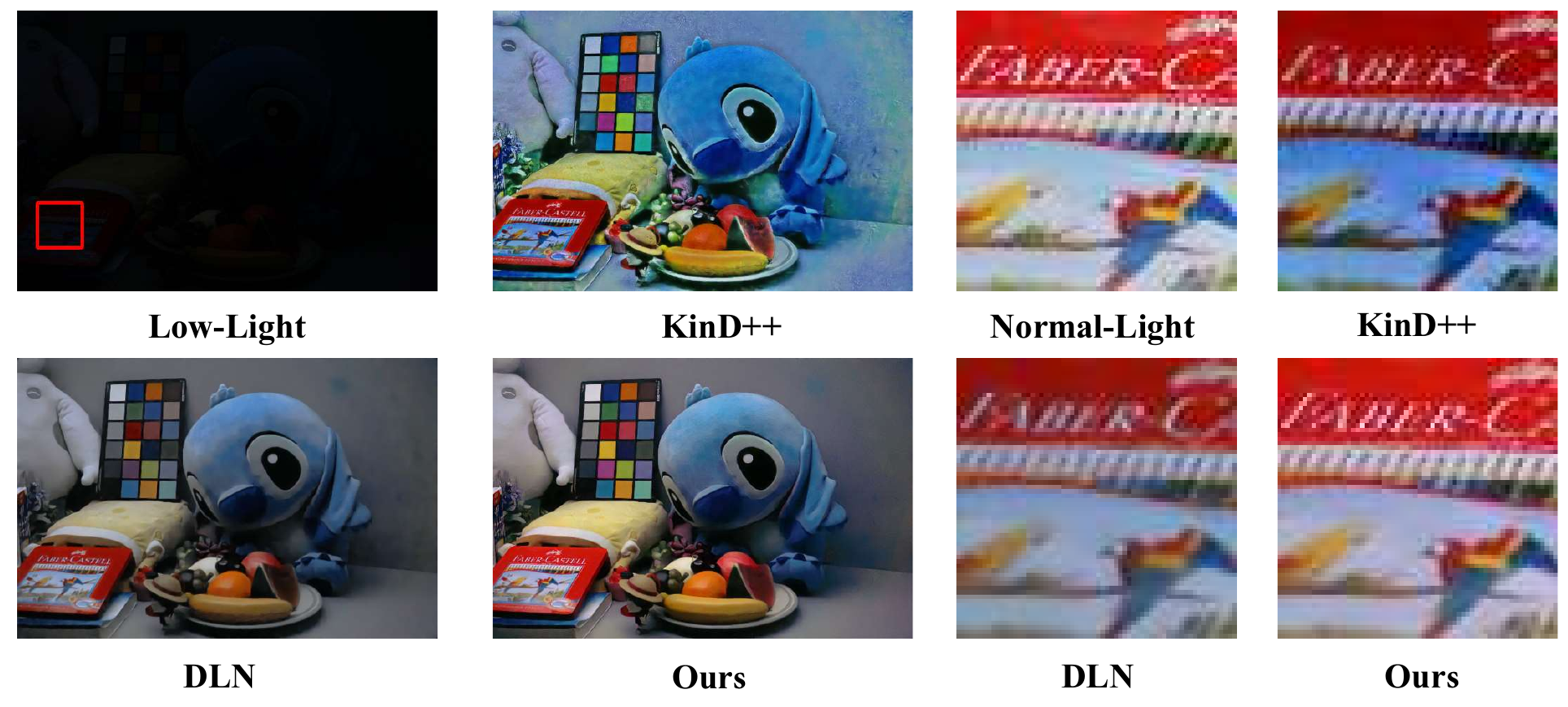}
\caption{Visual comparisons with two typical low-light image enhancement methods. The right side shows the zoom-in images of the selected small square area. (Image from LOL dataset). The proposed Surround achieves remarkable enhancement in brightness, color, and sharpness, while the other two methods generate image artifacts, otherwise low color saturation result.}
\label{fig_show_res}
\end{figure}

To address the above problem, we design a novel surround function, ASF (Adaptive Surround Function), to estimate the illumination map, which has the conventional surround function form and can be trained end to end. Base on the ASF, a reasonable and efficient network, SurroundNet, is constructed to bright the low-light image. We show some visual examples in Fig. \ref{fig_show_res} of our method compared with two SOTAs on one typical low-light image to illustrate its performance briefly. We can see that our method achieves visually promising results in terms of brightness, color, and sharpness. The contributions of our work are summarized as follows.
\begin{itemize}
\item \textbf{Light-weight low-light image enhancement}: The proposed SurroundNet achieves competitive performance against several state-of-the-art methods in various image quality measurements, with less than 150$K$ (80-98 percent size reduction compared to SOTAs) parameters.

\item \textbf{Adaptive Surround Function (ASF)}: We propose a new surround function, i.e., ASF, to estimate the illumination map which can be implemented by only two 1D convolution layers. Compared with stacking of 2D convolution layers and extra training in Retinex-CNNs \cite{wei2018deep}\cite{zhang2019kindling}, ASF has much fewer parameters and is easier to be optimized.

\item \textbf{Adaptive Retinex Block (ARBlock))}: We propose the ARBlock for both illumination adjustment and reflectance enhancement. The structure of ARBlock can be regarded as an extension of Single Scale Retinex (SSR) in features space. We present more discussion in Section \ref{arb}.

\item \textbf{Low-Exposure Denoiser (LED)}: As the brightness enhancement will usually amplify the noise, we further design a low-light denoise module to smooth the image before enhancement and use the synthetic low light image to supervise the training.

\end{itemize}
The rest of this paper is organized as follows: Section II summarizes the most related works. Section III describes the proposed SurroundNet. Section IV reports the comparison experiments conducted with SOTAs and analyzes the results. Section V gives a series of comparative ablation experiments. Finally, we conclude our study in Section VI.

\section{related works}

\subsection{Histogram based contrast enhancement}
As contrast degradation is the main character of low-light image, some simple and effective histogram-based methods like Histogram equalization (HE) \cite{pisano1998contrast, abdullah2007dynamic} are introduced to improve the visual perception of low-light image. However, those methods do not take the real brightness information into consideration, which may result in over- or under-enhancement\cite{guo2016lime}. Some advanced versions, like brightness preservation\cite{wang2005brightness} or contrast limitation\cite{reza2004realization}, can avoid the above issues to some extent. Nevertheless, histograms of different image part is very inconsistent \cite{wang2013naturalness} in serious non-uniform illumination cases, it makes the performance of those histogram-based methods not so attractive.

\subsection{Retinex based brightness enhancement}
Retinex theory indicates that a natural image can be decomposed into two parts, i.e., the reflectance and the illumination. In the Retinex framework, a low light image can be regarded as an image with low-intensity illumination part. The reflectance part only hinges on the scene itself, regardless of the lighting conditions. It is evident that the reflectance image can bring good visual performance in low-light image. Unfortunately, the decomposition of image is a high ill-posed problem\cite{wang2019progressive}. To settle the matter, surround function/operation is designed to estimate illumination part and then get reflectance sequentially. To be specific, we can get the illumination map by weighting surround pixels, and the type of weighting called the surround function. The surround functions have different form, such as $1/r^2$ function\cite{land1986alternative} and exponential form $e^{-|r|/c}$\cite{moore1991real}, where $r=\sqrt{x^2+y^2}$, $x,y$ are the coordinates and $c$ is a constant value. SSR is another early attempt, it serves the Gaussian filter $e^{-(r/\sigma)^2}$ as surround function to achieve the goal. For SSR, as mentioned above, it assumes the image can be decomposed into two parts:
\begin{equation}S = R \cdot I\end{equation}
where $S$ represents the image, $I$ and $R$ represent the illumination and reflectance. Accordingly, the light-free image $R$ can be estimated by the following form.
\begin{equation}R=logI-log[I*G]\end{equation}
where $G$ is the surround function defined as a Gaussian kernel:
\begin{equation}G(x,y)=Ke^{-\frac{x^2+y^2}{2\sigma^2}}\end{equation}
where $K$ is a weight to make $G$ comply with the following constraint.
\begin{equation}
\iint G(x,y) \, dx \, dy=1
\end{equation}
The logarithmic function converts division to subtraction and $*$ denotes the convolution operation. The original SSR research work \cite{jobson1997properties} investigates the logarithmic function and suggests a candidate formula as follows.
\begin{equation}R=logI-{log[I]*G}\end{equation}
Both of the two forms are valid in practice. In this paper, we choose the latter formula.

\begin{figure}[tp]
	\begin{center}
	  \subfigure[Surround Functions]{\includegraphics[width=0.225\textwidth]{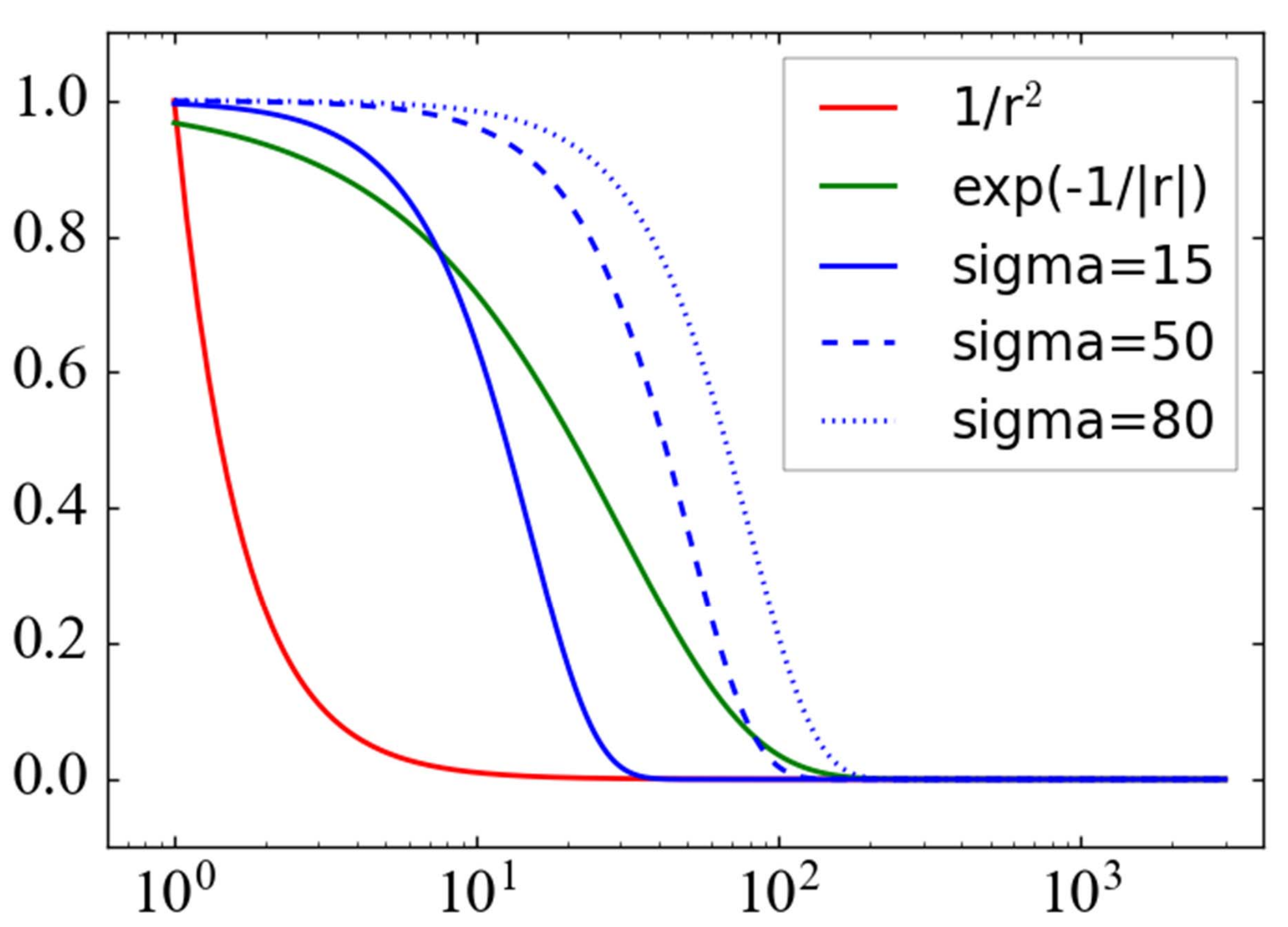}}\hspace{2mm}
	  \subfigure[$\sigma = 15$]{\includegraphics[width=0.24\textwidth]{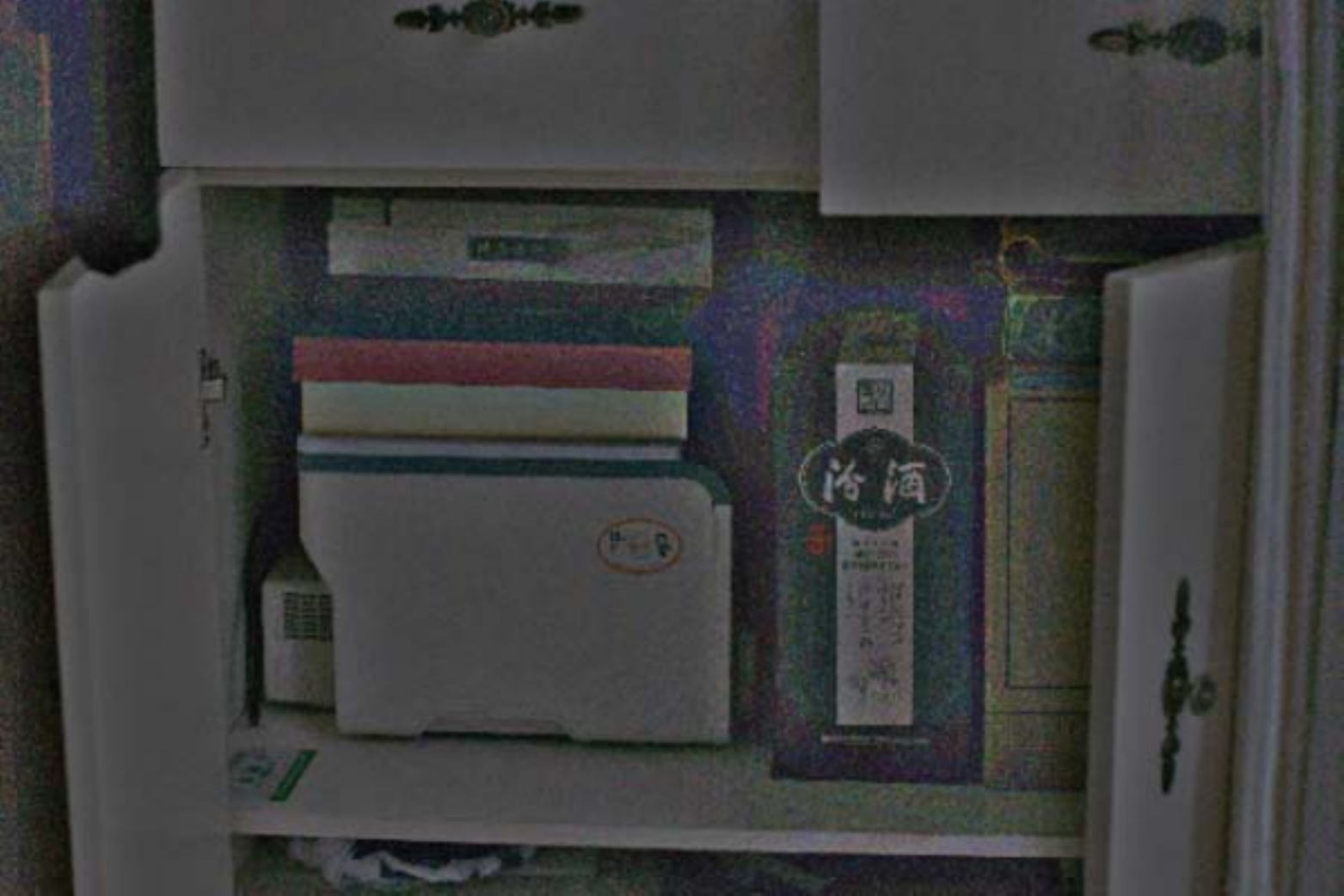}}
	  
	  \subfigure[$\sigma = 50$]{\includegraphics[width=0.24\textwidth]{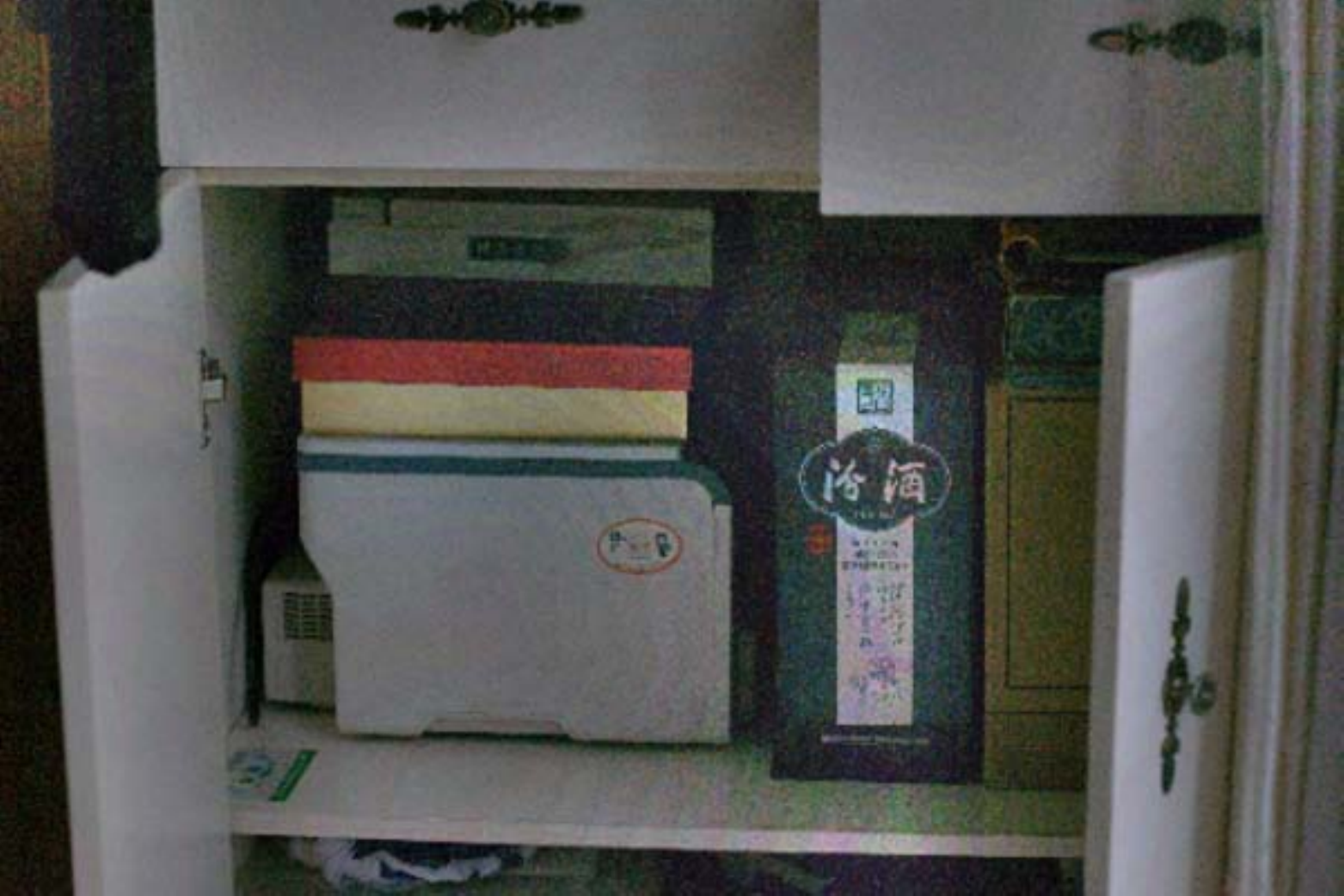}}
	  \subfigure[$\sigma = 80$]{\includegraphics[width=0.24\textwidth]{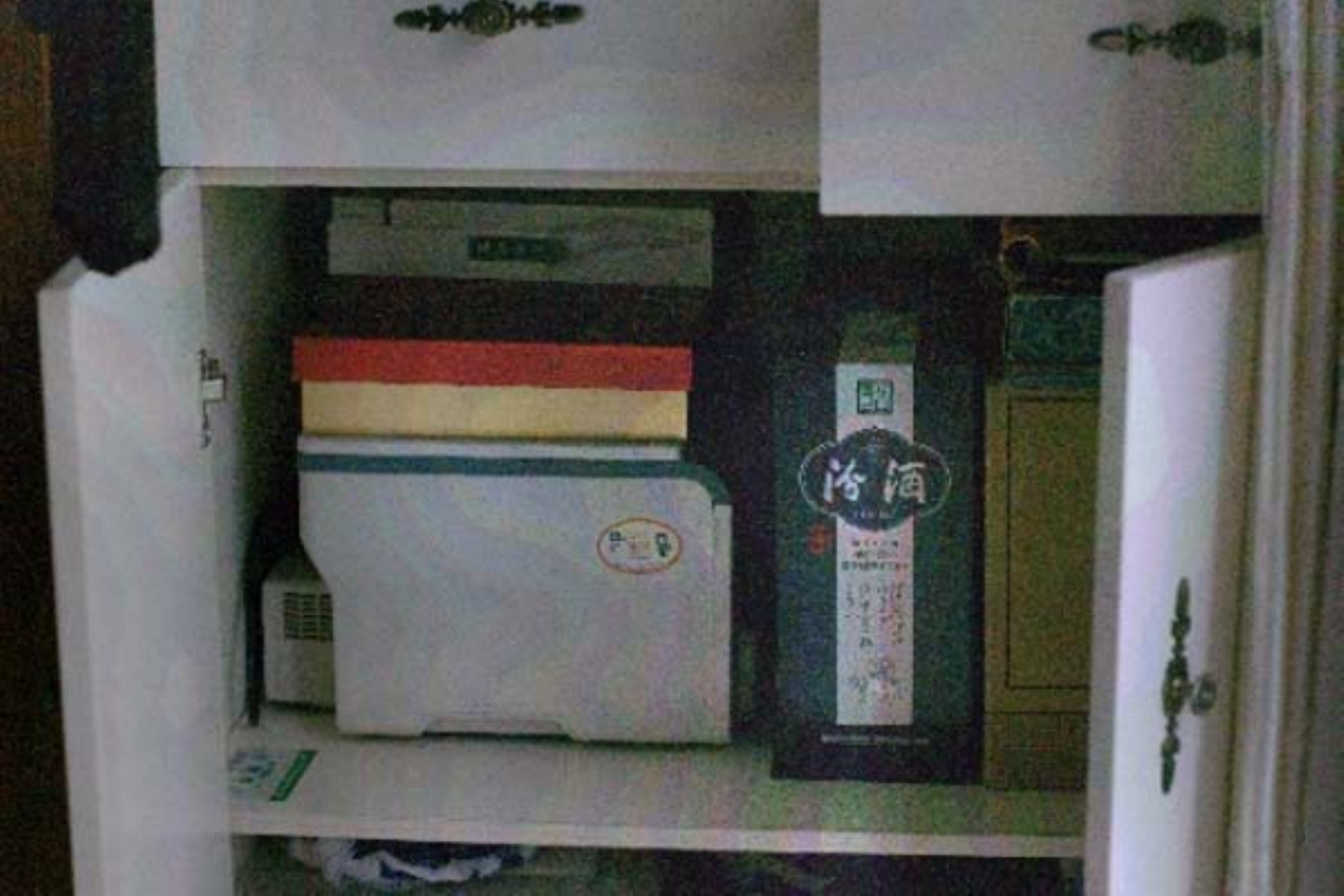}}
	\end{center}
	\caption{The top-left sub-figure is the comparison of three kind surround functions: inverse square $1/r^2$, exponential $e^{-|r|/c}$, and Gaussian $e^{-(r/\sigma)^2}$, where $c=30$, $\sigma=15, 50$ and $80$. And the rest three sub-figure are the Gaussian Retinex results with different $\sigma$.}
	\label{fig_SSR_sigma} %
	\vspace*{-1mm}
\end{figure}

Surround function is of great significance in Retinex-based methods. Some different surround functions can be found in Fig. 2. Although SSR declares the Gaussian filter surpasses others, the enhance result is sensitive to the variance parameter $\sigma$. Small $\sigma$ value means better dynamic range compression, and the opposite means natural tonal rendition\cite{jobson1997multiscale} as shown in Fig. \ref{fig_SSR_sigma}. Meanwhile, halo effect is another critical problem that has not been well solved. To provide both dynamic range compression and tonal rendition simultaneously and suppress halo phenomenon, Muti-Scale Retinex (MSR) uses different size Gaussian filters to obtain better illumination map and improve the visual quality. The multi-scale tactics work, but halo artifact still remains on edge where illumination change dramatically. Moreover, it is not easy to choose the size for different Gaussian filters. 
In this paper, we propose a novel Adaptive Surround Function (ASF), which can be regarded as a convolution kernel, to learn a general surround form. Details can be found in Section \ref{3b}.

\begin{figure*}[tp]
\centering
\includegraphics[width=1\linewidth]{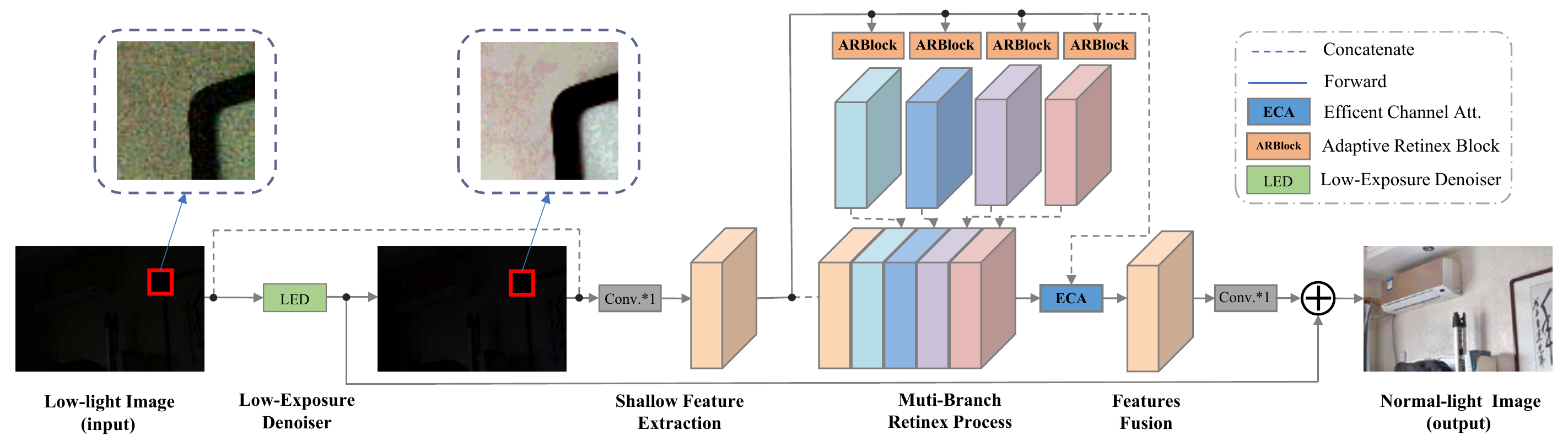}
\caption{Architecture of SorroundNet. The cube presents the feature maps of the corresponding operation which is denoted by colorful rectangle, the detailed exhibition of ARBlocks and ECA can be found in Section. 3}
\label{fig_sim}
\end{figure*}

\subsection{CNN based image enhancement }
LLNet \cite{lore2017llnet} attempts to enhance the low-light image with CNN (denoising auto-encoder). EnlightenGAN \cite{jiang2021enlightengan} employees the Generative Adversarial Networks (GANs) to transfer the low-light image to a normal image. Due to the gracefully mathematical foundation of Retinex theory, some work tries to combine the Retinex and CNNs. Retinex-Net \cite{wei2018deep} trains the sub-network (Decompose-Net) for decomposition and another one (Enhance-Net) for enhancing the illumination. MSR-net \cite{shen2017msr} investigates the relationship between the Multi-Scale Retinex and CNNs and uses the stack of convolutions to play an analogous role with Gaussian Surround Function. Progressive Retinex \cite{wang2019progressive} is more concerned with the signal-dependent noise problem. Two sub-networks, i.e., IM-Net and NM-Net, are designed to estimate illumination and noise level, respectively. Both the two networks mutually benefit from each other.

\subsection{Feature fusion}
Feature fusion has been demonstrated its ability on boosting the performance of CNNs, such as multi-scale feature fusion\cite{lin2017feature}\cite{szegedy2017inception} and channel attention mechanism.  For the low-light image enhancement task, MBLLEN\cite{lv2018mbllen} uses the multi-branch features to improve the image quality. And DLN\cite{wang2020lightening} have a similar multi-branch structure. Although the multi-branch processing broadens the feature channels significantly, it inevitably increases the number of parameters. Therefore, Efficient Channel Attention (ECA) \cite{qilong2020eca} is a good choice in reducing the parameters. It first takes the Global Average Pooling (GAP) \cite{lin2013network} to obtain aggregated features:

\begin{equation}
\begin{matrix}
g^c(\chi) = \frac{1}{W \times H}\sum_{x,y}^{W,H}\chi^{c}_{x,y}
\end{matrix}
\end{equation}
where $H,W,C$ are height, width and channel size, $\chi \in R^{H \times W \times C}$ is the output of prior convolution stage. GAP reduces the spatial dimension of feature maps. So it makes channel-wise processing possible. Channel Attention methods, such as Squeeze-and-Excitation (SE) \cite{hu2018squeeze}, apply a few dense layers on the output $g^c$ to get the weight of channels $F(g)$:
\begin{equation}
F(g) = \sigma(W_2{\rm ReLU}(W_1g))
\end{equation}
where ReLU is the Rectified Linear Unit and $\sigma$ represents the Sigmoid function, and $W_1$ and $W_2$ are the weight of two dense layers. The size of latent feature ${\rm ReLU}(W_1y)$ is usually less than the size of $g$. It benefits for reducing model complexity but injures the direct relationship between the $F(g)$ and $g$.
To avoid the issue, ECA uses the $1$D convolution to obtain the channel weight:
\begin{equation}
F(g) = \sigma({\rm C_{1D}^k}(g))
\end{equation}
where ${\rm C_{1D}}$ denotes 1D convolution, k is the kernel size of ${\rm C_{1D}}$. For 1D convolution, the kernel size is exactly equal to the number of parameters. It is obvious that ECA reduces the model complexity. In this paper, we modify original the ECA by using several $1$D convolutions to improve performance. Specially, the channel weight used in this paper turns into the following equation.
\begin{equation}
\label{eq9}
F(g) = \sigma({\rm C_{1D}^9}({\rm C_{1D}^9}(g)))
\end{equation}

\section{the proposed surroundNet}
\label{sec3}
\subsection{The architecture of SurroundNet}
Fig. 3 illustrates the structure of our proposed SurroundNet. We can see that the LED module firstly removes noise for the low-exposure condition. Then the shallow feature extraction (one $3\times3$ convolution) projects the denoised low-light image to feature space. After that, we design muti-branch ARblocks to enhance the image at various scales. The shallow feature and enhanced results will be combined into the ECA module. Finally, another $3\times3$ convolution layer will fuse the features from ECA and produce the lightened result. The ARBlock and LED modules achieve the model reduction in network size. Even the ECA module slightly increases the number of parameters by utilizing larger kernel and several 1D convolution layers (as shown in equation \ref{eq9}), it is still lightweight.

\subsection{Adaptive Surround Function}
\label{3b}
Previous researches about the surround function \cite{jobson1997properties} exhibit that Gaussian Surround performs better than the $1/r2$ Surround and the Exponential Surround. However, the selection of Gaussian parameters is empirical and can not handle the complex illumination environment. In this work, we design a new Adaptive Surround Function (ASF) whose form can be learned in a data-driven way and does not stick to a certain surround form. Specifically, it is implemented by one convolution layer, which means our ASF is a special convolution kernel.

\begin{figure}[th]
\centering
\includegraphics[width=1\linewidth]{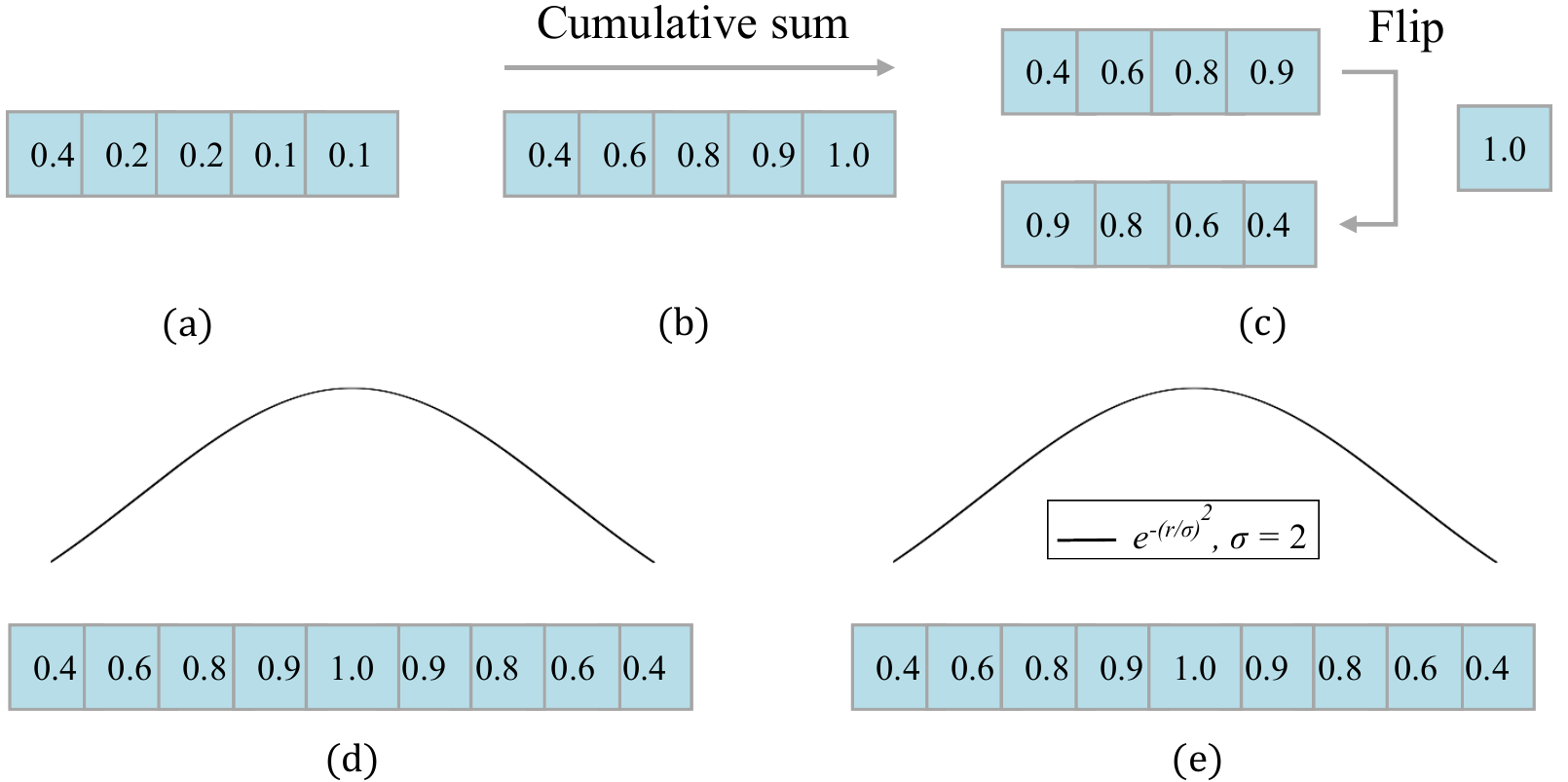}
\caption{The visualization of building the ASF convolution kernel. In this toy example, $K$ is set to 5 and the normalization step is omitted.}
\label{fig_ASF}
\end{figure}

Fig. \ref{fig_ASF} visualizes the procedure of constructing the ASF (convolution kernel) as a toy example. At first, we built a $1$D convolution kernel $x$ with $K$ parameters to learn as the input (e.g., Fig. \ref{fig_ASF}(a)). Then we cumulative sum $x$ to guarantee it can be monotonically increased as shown in Fig. \ref{fig_ASF}(b). Fig. \ref{fig_ASF}(c) utilizes the mirror flip $\dot{x}$ to get $\ddot{x}$ except the last element. After that, we concatenate $\dot{x}$ and $\ddot{x}$ to get a symmetrical and larger receptive field kernel $K_{ASF1D}$ as shown in Fig. \ref{fig_ASF}(d). Finally, we normalize $K_{ASF1D}$ with Equ. 4. For intuitive comparison, Fig. \ref{fig_ASF}(e) shows a Gaussian surround which has the same weights with our ASF. So Gaussian Surround can be considered as a special case of our ASF. In fact, we can get various kinds of Surround Functions (e.g., inverse square and exponential) by set $x$ at an appropriate value. In our experiments, $x$ is learnable and we initialize $x$ to an all-one vector.

Moreover, the details of the construction algorithm of 1D-ASF can be found in Algorithm 1. The constructing process shows that a $K$ size ASF possesses $(2*K-1)$ receptive field. It is spatially symmetrical and has the most response on the center pixel. All the operations introduced above are derivable and can be optimized end to end.
\begin{algorithm}[h]
\caption{1D-ASF kernel construction}
\hspace*{0.02in} {\bf Input:} 
1D-vector $x$ with size $K$\\
\hspace*{0.02in} {\bf Output:} 
Surround Function $K_{ASF1D}$
\begin{algorithmic}[1] 
\Statex // Processing the vector with cumulative sum operation. 
\State $\dot{x}$ = Cumsum(Abs(x))
\Statex // Mirror and concatenate. 
\State $\ddot{x}$ = Fliplr($\dot{x}$[:-1]) 
\State s = Concat($\dot{x}$,$\ddot{x}$)
\Statex // Normalization 
\State $K_{ASF1D}$ = s / Sum(s) 
\State \Return $K_{ASF1D}$
\end{algorithmic}
\end{algorithm}

The 1D-ASF can be extended to a 2D form by multiplying the transpose of itself (i.e., Eq.\ref{eqasfV2}).

\begin{equation}
\label{eqasfV2}
K_{ASF2D} = (K_{ASF1D} \cdot K^{\mathrm{T}}_{ASF1D})
\end{equation}
where $K_{ASF2D}$ is 2D-ASF kernel and $K_{ASF1D}$ defines 1D-ASF kernel. Moreover, to be effective, 2D-Gaussian kernel could be divided into two 1D-Gaussian kernels and can be described mathematically as follow:
 
\begin{equation}
I*G_{2D} = I*G_{1D}*G^{\mathrm{T}}_{1D}
\end{equation}
where $I$ is the image, $G_{2D}$ is a 2D-Gaussian kernel, $G_{1D}$ and $G_{1D}$ are two 1D-Gaussian kernel, and $*$ means convolution operation.

Obviously, the combination of two separable 1D-kernels has much fewer parameters than a large 2D-kernel. Therefore, we formalize our $2$D ASF into a lightweight module as shown in Eq. \ref{eqasf}.
\begin{equation}
\label{eqasf}
I*K_{ASF2D} = I*K_{ASF1D}*K^{\mathrm{T}}_{ASF1D}
\end{equation}

\subsection{Adaptive Retinex Block}
\label{arb}
In the previous Retinex-based works like SSR, surround function should work on the image space (gray or RGB). However, in this work, we expand the scope of surround function from image space to the feature space, which can be obtained by pre-handing (stack of several convolution layers). Such an assumption is reasonable because the low-pass trait of ASF (just like other surround function) can extract the low-frequency features. And frequence separation strategy has been demonstrated its effectiveness in improving the performance of CNNs \cite{zhou2020guided}. More discussion and experiments can be found in Section. IV.
\begin{figure}[ht]
\centering
\includegraphics[width=1\linewidth]{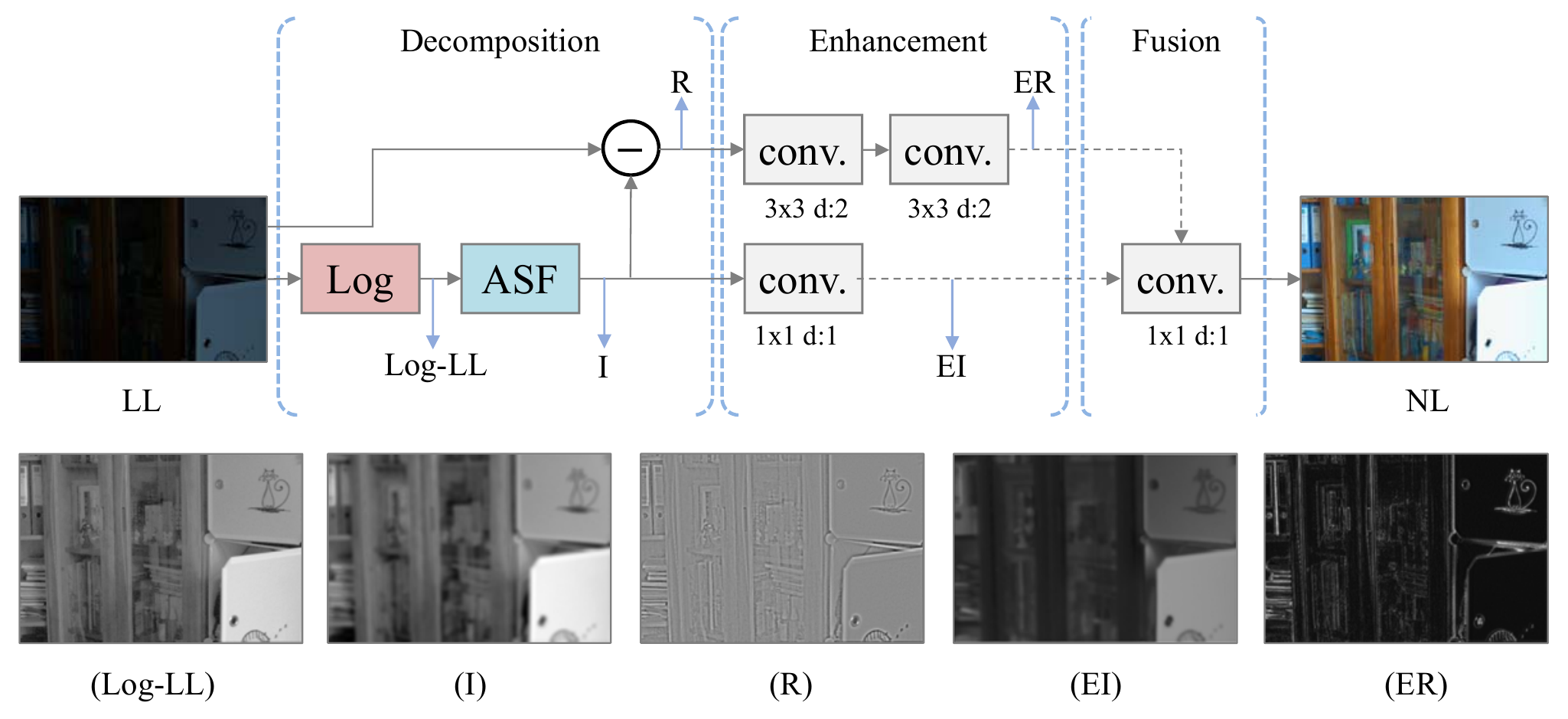}
\caption{The architecture of ARBlock. The five images as shown below are the visualized results by averaging the channel of the feature maps.}
\label{fig_ARB}
\end{figure}
Adaptive Retinex Block (ARBlock) is designed for both illumination adjustment and reflectance enhancement, which subsists of three steps. Fig. 5 shows the architecture of our ARBlock. At first, we apply a logarithmic transformation to the input low-light image (LL). The illumination (I) can be obtained by convolving LL and ASF. And the reflectance (R) is the subtraction result of LL and I. We can see that the logarithmic transformation lights the dark part. I represents the smooth and low-frequency part of the image, and R is about the edges. In the enhancement step, we enhance the I and R respectively. We use an illumination adjustment operation (i.e., one $3\times3$ convolution layer) on I to promote the brightness and reduce the halo effect of the surround function. And we take a reflectance enhancement operation (i.e., two $3\times3$ convolution layers with dilation 2) on R to correct the color and smooth the noise. Different with conventional Retinex, our work aims to get normal-lighted image (NL) rather than only reflectance image. Therefore, the last step is to fusion the enhanced I (EI) and R (ER). We use one $1\times1$ convolution to fusion the EI and ER rather than add them. We adopt 4 ARBlocks in our experiments, and the $K$ of ASF in different ARBlocks are set to 3,7,11, and 15, respectively.

\subsection{Low-Exposure Denoiser}
\label{led}
\begin{figure}[ht]
\centering
\includegraphics[width=1\linewidth]{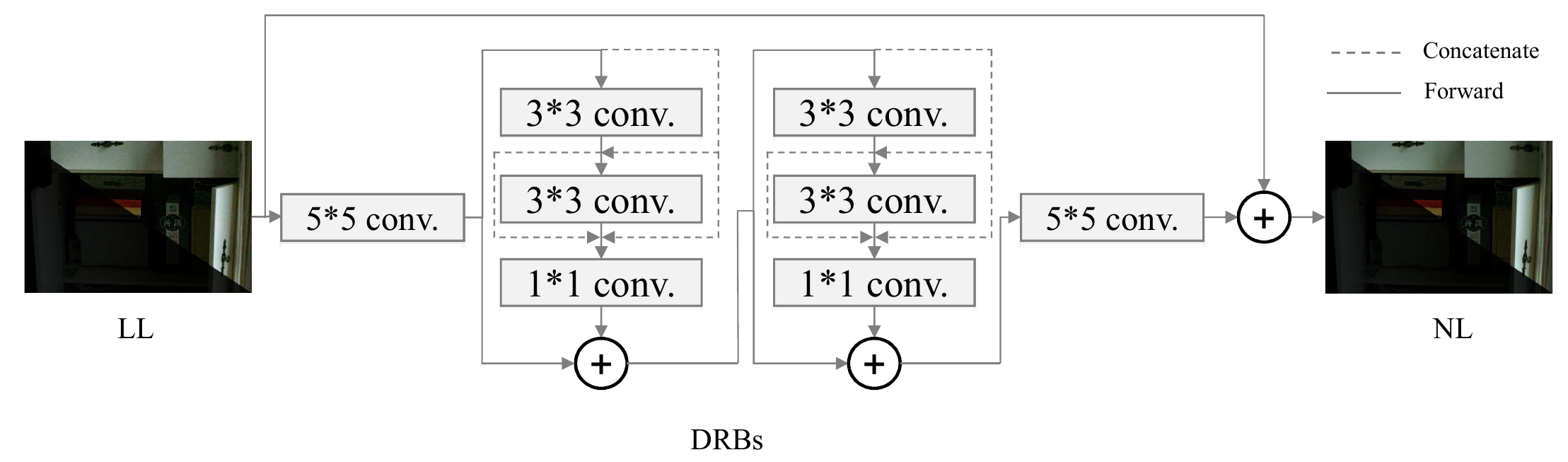}
\caption{The architecture of LED. Here, we omit the ReLU layer.}
\label{fig_LED}
\end{figure}

Many factors can cause the noise of the image, such as dark current and electronics shot in camera imaging \cite{liu2007automatic, tsin2001statistical}. In the enhancement procedure, the noise in the image will be amplified \cite{wang2019progressive}. Therefore, we design the Low-Exposure Denoiser (LED) module to remove the noise before enhancement and use synthetic noise-free dark image to supervise the module learning. The structure of LED is shown in Fig. \ref{fig_LED}. Densely connection\cite{huang2017densely} and residual connection\cite{he2016deep} have been proved their effective in image restoration task \cite{park2019densely}\cite{tong2017image}. Residual Dense Network\cite{zhang2020residual} proposes residual dense block (RDB) to combine both of them. The RDB can reuse the low-level feature by densely connected convolutional layers, and its local and global feature fusion module makes training stabilized. Inspired by \cite{zhang2020residual}, our LED employs RDB module as component. As shown in Fig. \ref{fig_LED}, the first $5\times5$ convolution layer extracts the shallow features. Then we apply two RDBs on the shallow feature map. The last layer is also a $5\times5$ convolution which fuses the features and produces clear dark image. In this paper, we use $5\times5$ kernel convolution rather than two $3\times3$ convolution to extract shallow features and do not adopt batch normalization\cite{ioffe2015batch}. The reason of the former operation is that the input image only has $3$ channel, so large kernel in first and last layer can obtain large receptive field with few parameters. The latter is because we found batch normalization will lead to convergence instability.

\subsection{Loss function}
Our SurroundNet optimizes the parameters in a fully supervised way. For every dark image, there will be a light image as its training target. Zhao et al. \cite{zhao2016loss} investigate various loss functions in the image restoration task and suggests the combination of L1 loss and MS-SSIM loss \cite{wang2003multiscale}. MIRNet\cite{zamir2020learning} employs Charbonnier loss\cite{charbonnier1994two} for various different image restore tasks like denoising, super-resolution, enhancement and achieves state of art for all of them. DLN studies different loss function performances in low-light enhancement task and finds that the combination of SSIM loss\cite{wang2004image} and TV Loss has the best performance. Perceptual loss is also used widely in low-light image enhancement\cite{lv2019attention}\cite{lv2018mbllen}. In this paper, inspired by the above investigations, we use a combination of three loss functions (i.e., SSIM, Charbonnier and DISTS perceptual loss\cite{ding2020image}) for the whole network :
\begin{equation}
L_{t}= L_{h} + L_{l}
\end{equation}
\begin{equation}
L_{h}= L_{ssim}(o,n) + L_{char}(o,n) + L_{dists}(o,n) 
\end{equation}
\begin{equation}
L_{l}= L_{ssim}(ol,cl) + L_{char}(ol,cl) + L_{dists}(ol,cl) 
\end{equation}
where $o$ defines the output image, $n$ defines the normal-light image, $cl$ is the clear noisy-free low-light image, and $ol$ is the output of LED.

We adopt SSIM loss for denoising and structure reservation. It is defined as follows. 
\begin{equation}
L_{ssim}(x,y)= 1 - \frac{2\mu_x\mu_y+c_1}{\mu_{x}^{2}+\mu_{y}^{2}+c_1}\cdot\frac{2\theta_{xy}+c_2}{\theta_{x}^{2}+\theta_{y}^{2}+c_2}
\end{equation}
where $\mu_x$ and $\mu_y$ are the mean values of images, $\theta_{x}$ and $\theta_{y}$ are the variances, and $\theta_{xy}$ is the covariance. There are also two hyper-parameter $c_1$ and $c_2$ to avoid dividing by zero, both of them are set as suggested in \cite{wang2020lightening}.

SSIM loss has been proved that can cause shifts of colors \cite{zhao2016loss}. Fortunately, the $L1$ loss has the ability to restore colors and luminance. Therefore, it is better to combine them and capture the best performance. However, for low-light images, color degradation is inevitable and hard to restore. The outliers of the result may be harmful to network optimization. Therefore, we use the robust Charbonnier loss to replace $L1$, which is defined as:
\begin{equation}
L_{char}(x,y) = \sqrt{||x-y||^{2}+\epsilon^{2}}
\end{equation}
where $\epsilon$ is a constant and we set it to $10^{-3}$.

We also take DISTS perceptual loss to further improve the visual quality. It first transforms the image to “perceptual” representation by using pre-trained VGG19 \cite{simonyan2014very} (defined as $f(\cdot)$). Then the “perceptual” representation $f(\cdot)$ will be handled by texture measurement ($l(\cdot)$) and structure measurement ($s(\cdot)$) as follows.
\begin{equation}
l(f(x),f(y)) = \frac{2\mu_f(x)\mu_f(x)+d_1}{\mu_{f(x)}^{2}+\mu_{f(x)}^{2}+d_1}
\end{equation}
\begin{equation}
s(f(x),f(y)) = \frac{2\theta_{f(x)f(y)}+d_2}{\theta_{f(x)}^{2}+\theta_{f(y)}^{2}+d_2}
\end{equation}
where $d_1$ and $d_2$ are constant values to avoid numerical instability, $\mu_f(x)$ and $\mu_f(y)$ are the mean values of perceptual representations, $\theta_{f(x)}$ and $\theta_{f(y)}$ are the variances and $\theta_{f(x)f(y)}$ is the covariance.

The integrated DISTS measurement is a weighted sum of texture and structure, so DISTS Loss can be defined as:
\begin{equation}
L_{dists}(x,y) = 1 - (\zeta l(f(x),f(y)) + \eta s(f(x),f(y)))
\end{equation}
where $\zeta$ and $\eta$ are the positive learnable parameters, which are set as suggested in \cite{piq}.

\section{Experiments}

\subsection{Real world dataset}

Low-Light (LOL) dataset\cite{wei2018deep} is a publicly available dark-light paired images dataset in the real sense. The low-light images are collected by changing exposure time and ISO. It contains 500 images in total, we use 485 images of them for training, and the rest for evaluation as suggested by \cite{wei2018deep}.

\begin{figure}[h]
\centering
\includegraphics[width=1\linewidth]{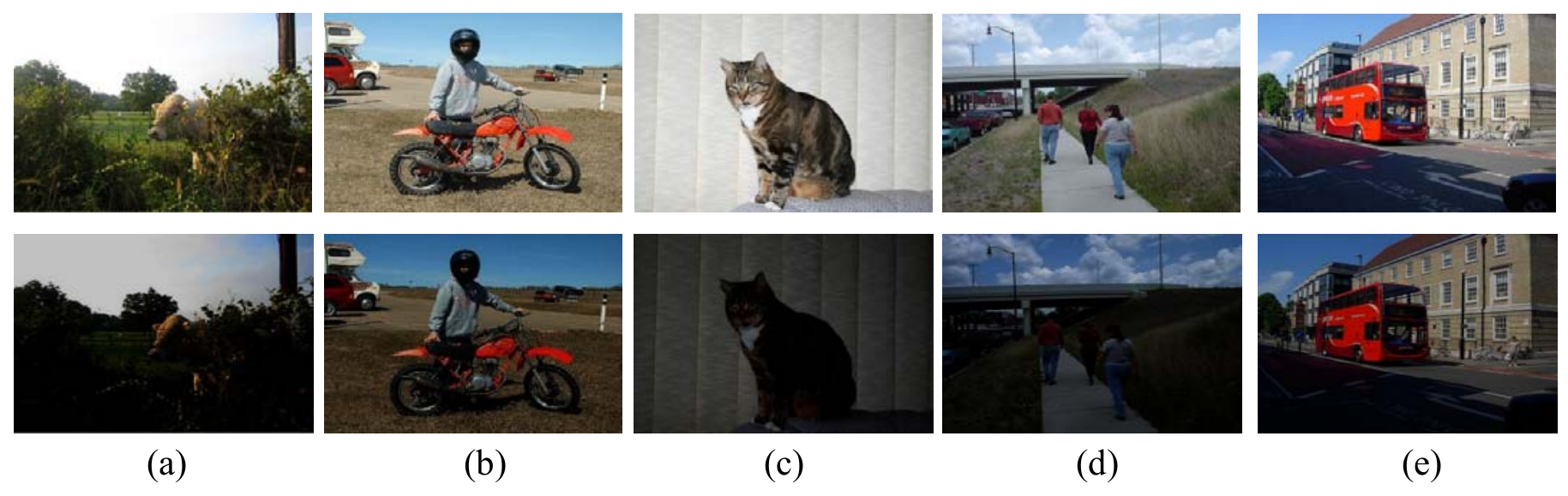}
\caption{Some synthetic low-light images and their normal-light ones in PASFAL VOC 2007 dataset.}
\label{fig_synData}
\end{figure}

\begin{figure*}[h]
\centering
\includegraphics[width=1\linewidth]{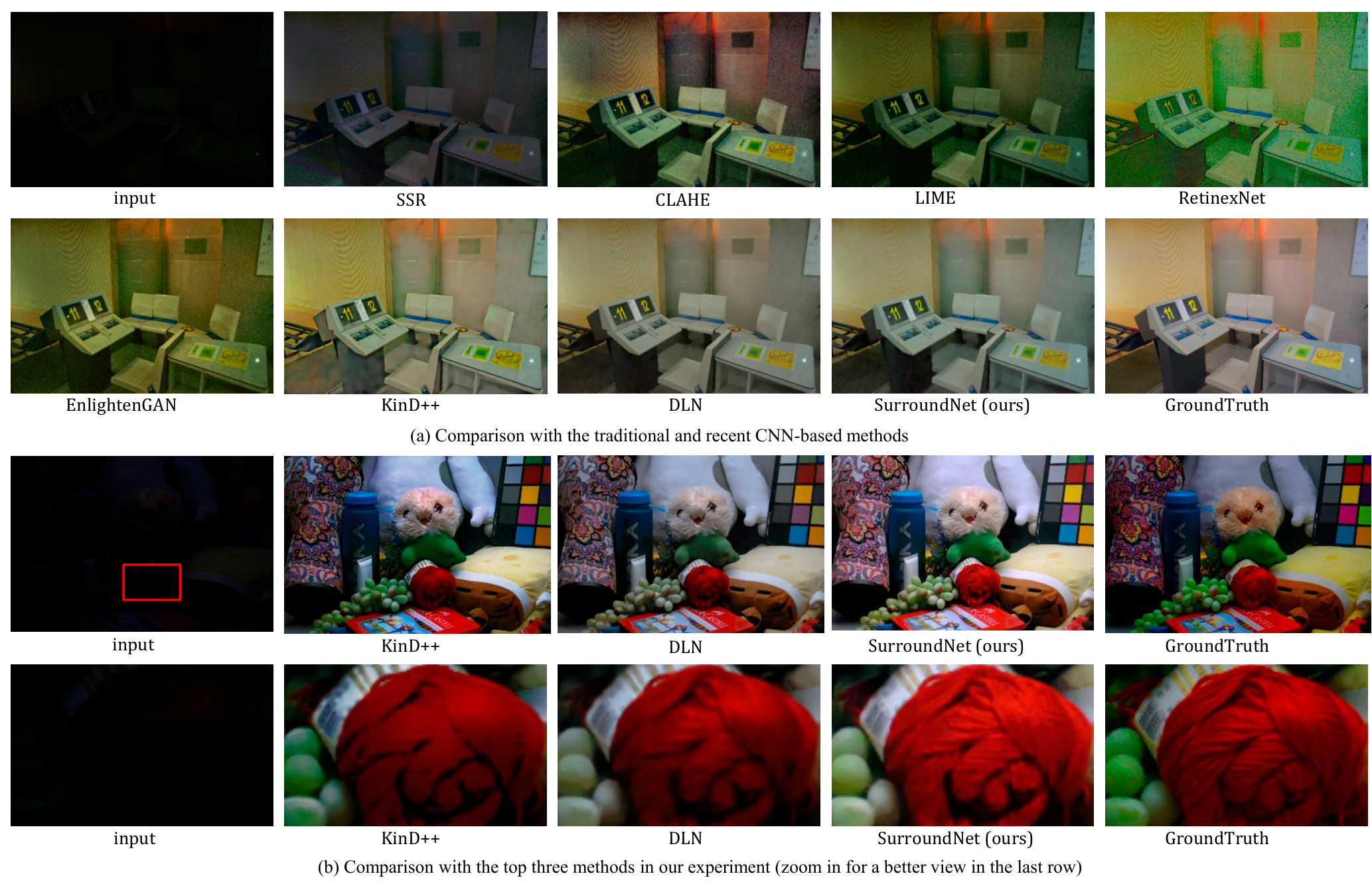}
\caption{Enhance results of different algorithms on LOL dataset }
\label{fig_compare}
\end{figure*}

\subsubsection{Synthetic dataset}
\label{syndata}
For low-light image enhancement task, it is hard to acquire dark-light pairs for real scene at the same time. Even the LOL dataset \cite{wei2018deep} provides a real-world image dataset, we still need a huge amount of images for training the deep networks. Lv et al. \cite{lv2019attention} gave an effective synthesize low-light image method from normal exposed image. The idea is that a low-light image can be simulated through the combination of linear and gamma transformation. So the math formulation can be shown as follow.
\begin{equation}
I_{low}^{(R,G,B)} = \beta \times (\alpha \times I_{high}^{(R,G,B)})^\gamma
\end{equation}
where $\beta$, $\gamma$, and $\alpha$ are randomly sampled from uniform distribution. When the symbols $\gamma$ is greater than $1.0$, the dynamic range of dark region in image will be compressed. The $\alpha$ and $\beta$ will control the max value of bright part. This method will darken the normal light image which could achieve a close visual perception from the real low-light image. We faithfully follow the parameter settings in \cite{lv2019attention}: $\beta \sim U(0.5,1)$, $\alpha \sim U(0.9,1)$, $\gamma \sim U(1.5,5)$. The normal light image dataset should have both good visual quality and diversiform sense. We choose the PASCAL VOC 2007 dataset which has been widely used in various vision tasks like object detection\cite{ren2015faster} and semantic segmentation\cite{girshick2014rich}. We use all the split sets (training, validation, testing) in the VOC2007 dataset which contains 9,963 images. In addition, we present some examples of the synthetic images in Fig. \ref{fig_synData}.

\subsubsection{Noise-free low-exposure data}
As mentioned in Section \ref{led}, we need noise-free low-light images to supervise the LED module training. However, noise in low-light image is signal-dependent \cite{wang2019progressive} and difficult to be simulated. In this work, we synthesize the noise-free low-light images on real dataset (LOL). LOL contains clear normal-light image and noisy low-light image pairs. Inspired by the above synthetic method, we can generate the noise-free low-light image from the normal light image in LOL. It is worth mentioning that the three parameters, i.e., $\beta$, $\alpha$ and $\gamma$, should be set adaptively to reduce the difference between synthetic dark image and real dark image as far as possible. In this work, we treat it as a least-squares optimization problem as shown below.
\begin{equation}
Argmin_{\alpha,\beta,\sigma} (\sum_{i=1}^{N}(I_{low,i}^{(R,G,B)}-\beta \times (\alpha \times I_{{high,i}}^{(R,G,B)})^\gamma))
\end{equation}
where $i$ is the pixel index and $N$ is the number of pixels. For each dark-light image pair in the LOL dataset, we use Levenberg-Marquardt algorithm\cite{more1978levenberg} to optimize the three parameters. We will use the optimized parameters to synthesise the required noise-free low-light images. As for synthetic dataset, our synthetic process (Section \ref{syndata}) does not involve any synthetic noise, therefore, the supervision target of LED is the dark image itself.

\subsection{Training settings}
Except ASF module (\ref{3b}), all weights of our SurroundNet are initialized randomly. Adam optimizer is adopted to optimize the whole network. The learning rate is set to 0.001, and the momentum is 0.9. In every mini-batch, we randomly crop 32 dark-light patch pairs, each of which has two paired 128*128 images. We set epoch to 100 on synthetic dataset, and fine-tune epoch is set to 3500 on LOL dataset. For ablation experiments, all the training epochs are set to 100 unless otherwise stated. All models are trained on the platform with Nvidia GTX 2080Ti GPU. 

Our code is open source and can be downloaded from the following GitHub repository: https://github.com/ouc-ocean-group/SurroundNet.

\subsection{Compare with State-of-the-Art methods}
We compare our SurroundNet with five the-state-of-art low-light enhancement methods, i.e., LIME (TIP, 2017) \cite{guo2016lime}, RetinexNet (BMVC, 2018) \cite{wei2018deep}, EnlightenGan (TIP, 2021) \cite{jiang2021enlightengan}, kinD++ (IJCV, 2021) \cite{zhang2021beyond} and DLN (TIP, 2020) \cite{wang2020lightening}. LIME is one representative conventional method via illumination map estimation, and the rest four methods employ the deep CNN for image enhancement. Specifically, LIME utilizes the maximum value of RGB to roughly estimate the initial illumination and use structure-aware smoothing to get final result \cite{guo2016lime}. RetinexNet introduces relectance consistent loss to learn illumination in a data-driven way \cite{wei2018deep}. EnlightenGan employs adversarial learning to enhance low-light image without paired training data \cite{jiang2021enlightengan}. KinD++ decomposes image into illumination and reflectance parts and adopts multi-scale illumination attention module to restrain visual defects \cite{zhang2021beyond}. DLN brings Back-Projection concept into low-light enhancement from super-resolution tasks \cite{wang2020lightening}. 

We adopt four quality metrics (PSNR, SSIM, NIQE\cite{mittal2012no}, LPIPS\cite{zhang2018unreasonable}) to evaluate those methods. NIQE is a no-reference metric on the strength of natural scenes statistics. LPIPS measures the visual quality by computing the weighted $L_2$ distance on feature space.

Table \ref{table1} shows the comparison results with the five methods. From the results, we can see that our SurroundNet is superior to the others with much fewer parameters. For example, we exceed the second best method, i.e., DLN, by $0.87$ on PSNR and $0.05$ on SSIM but only use about $20$ percent parameters of DLN. It certifies that our SurroundNet is very effective in low-light image enhancement. The best performance on PSNR and SSIM clarifies that our method can hold both brightness enhancement and noise suppression. Meanwhile, the LIPIS index certifies that our method achieves the best visual perception. KinD++ is the improved version of KinD \cite{zhang2019kindling} and achieves competitive results. It is worth mentioning that the number of parameters for kinD++ and EnlightenGan is more than ten times higher than our method. This is because both of them employ Unet-like \cite{ronneberger2015u} structure with several down-sampling modules, which makes them have to spend a large amount of parameters on restoring the lost spatial information. In contrast, DLN and our SurroundNet methods perform on high-resolution image and feature maps. Therefore, they do not have the trouble of losing spatial information.
\begin{table}[bhp]
\centering
\caption{Comparison with state-of-the-art methods (\red{red}: Best; \blue{blue}: the $2^{RD}$ Best; \green{grenen}: the $3^{RD}$ Best)}
\label{table1}
\begin{tabular}{cccccc}
\hline
module &PSNR$\uparrow$ &SSIM$\uparrow$ &NIQE$\downarrow$ &LPIPS$\downarrow$ &Params$\downarrow$\\ \hline
LIME &14.02 &0.513 &8.089 &0.391 &/ \\ \hline
RetinexNet &16.77 &1.699 &8.878 &0.467 &\blue{440k}\\ \hline
EnlightenGan &17.48 &0.651 &\blue{4.686} &0.390 &8643k\\ \hline
kinD++ &\green{21.80} &\green{0.834} &5.118 &\green{0.289} &8274k\\ \hline
DLN &\blue{21.94} &\blue{0.848} &\green{4.882} &\blue{0.259} &\green{700k}\\ \hline
SurroundNet &\red{22.81} &\red{0.853} &\red{4.384} &\red{0.190} &\red{137k}\\ \hline
\end{tabular}
\end{table}

Fig. \ref{fig_compare} gives some visualized results. The comparison includes three conventional methods (SSR \cite{jobson1997properties}, CLAHE \cite{pisano1998contrast} and LIME) and four above mentioned CNN-based methods. SSR \cite{jobson1997properties} explores different surround functions and recommends the Gaussian Surround to estimate illumination. The CLAHE is a histogram-based method which can suppress over-enhancement by limiting the overstretch of histogram \cite{pisano1998contrast}. From Fig. \ref{fig_compare}, we can find that CNN-based methods exhibit better performance than the conventional ones. And among all the five CNN-based methods, our SurroundNet and DLN achieve the most natural illumination adjustment. Moreover, our method exceeds DLN in image structure preservation. Sub-figure (b) also displays the zoom-in results of the three best CNN-based methods. We can clearly see that SurroundNet can restore the texture of red string ball while the other two methods give the blur results.

\begin{figure*}[htp]
\centering
\includegraphics[width=0.95\linewidth]{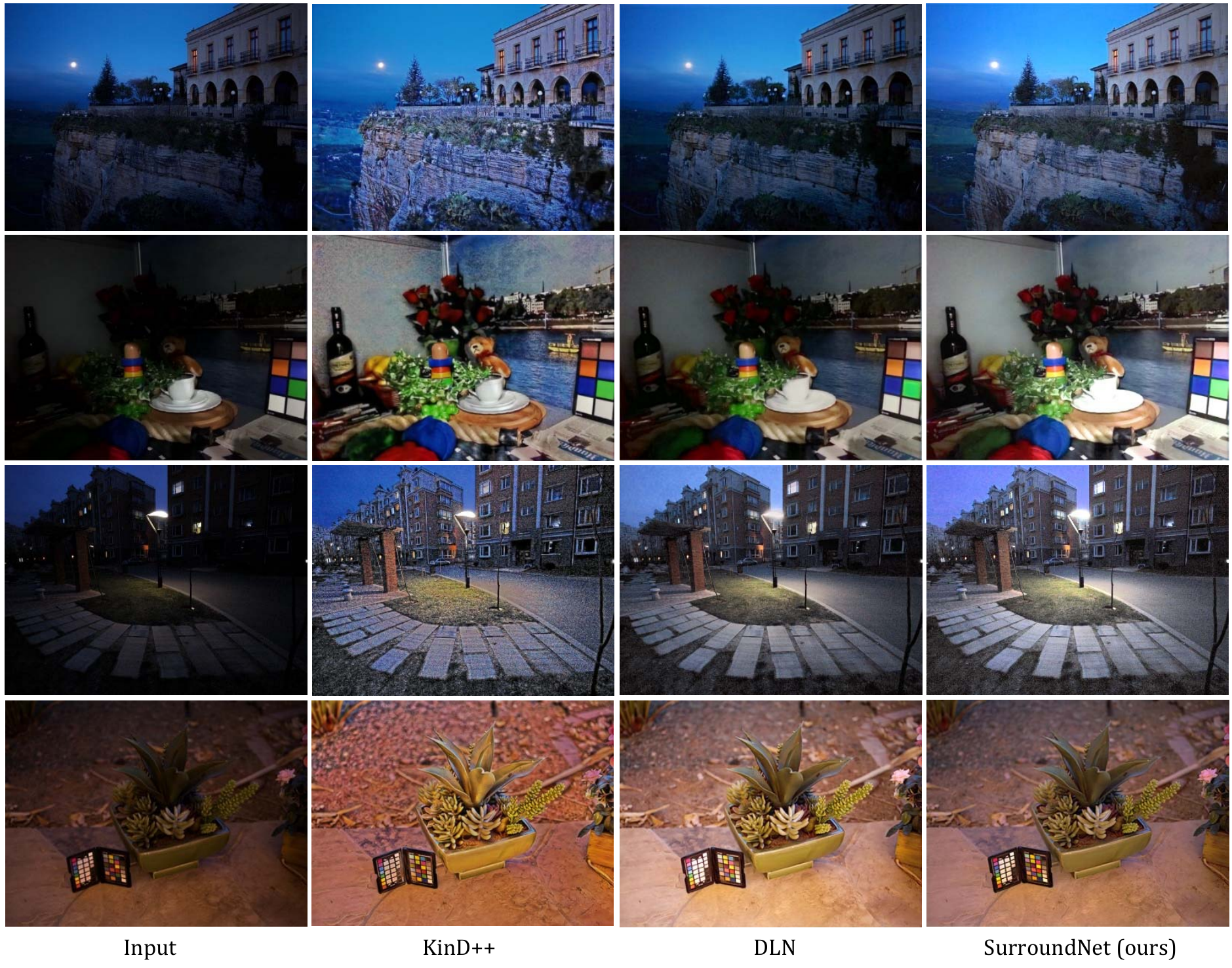}
\caption{Some comparison examples on the LIME dataset.}
\label{fig_compareLIME}
\end{figure*}

\begin{figure*}[htp]
\centering
\includegraphics[width=0.95\linewidth]{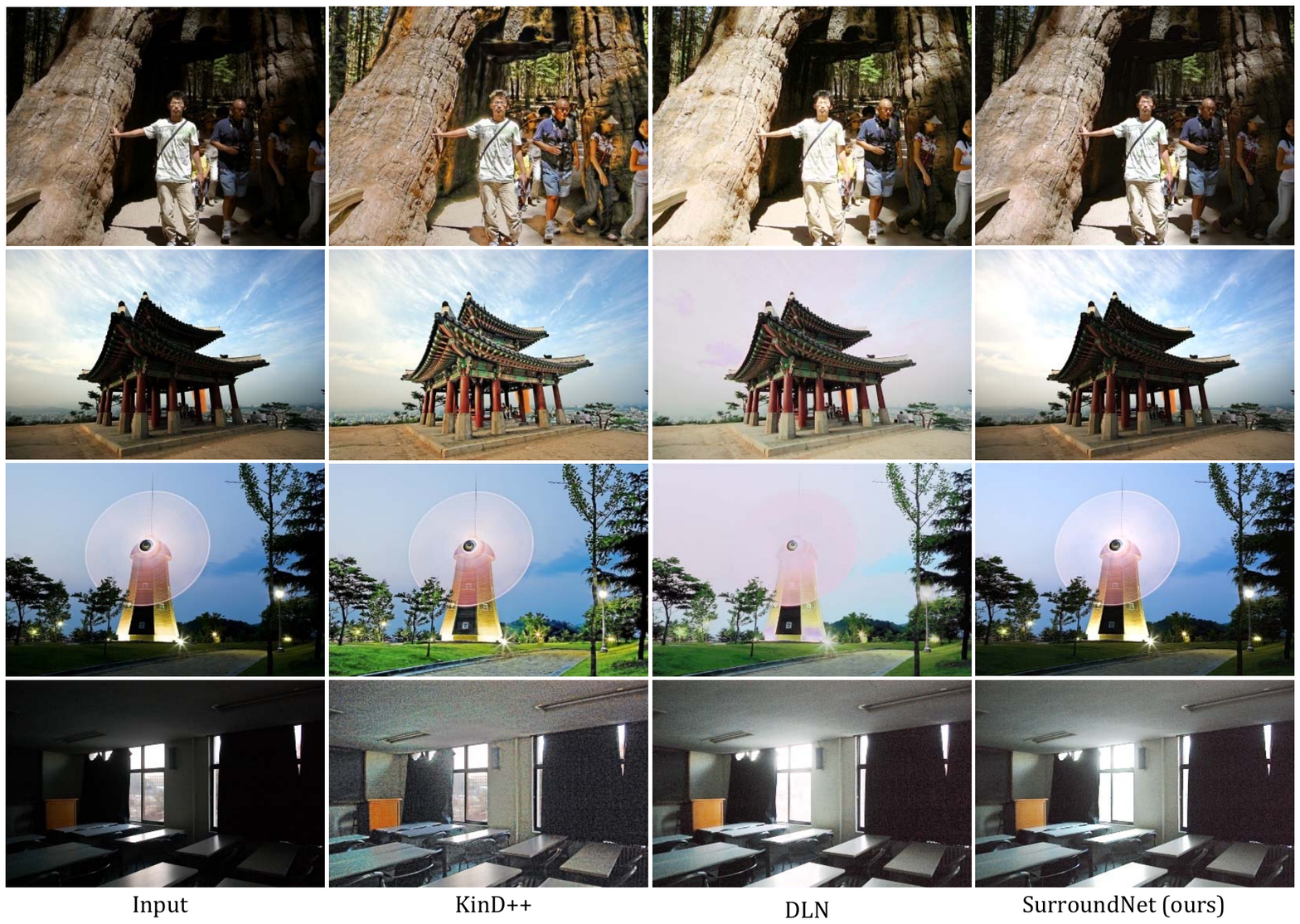}
\caption{Some comparison examples on the DICM dataset.}
\label{fig_compareDICM}
\end{figure*}

To verify the generalization ability of our SurroundNet, we further evaluate it on two widely-adopted datasets: LIME (10 images) \cite{guo2016lime}, and DICM (64 images) \cite{6615961}. Both of these two datasets are unpaired low/normal light data. Some samples of them are shown in  Fig. \ref{fig_compareLIME} and Fig. \ref{fig_compareDICM}. We observe that SurroundNet offers similarity denoising ability with DLN but more proper light adjustment, especially in some bright regions like the sky. We also find our model performs the best in the indoor scene like the second image in Fig. \ref{fig_compareLIME} and last image in Fig. \ref{fig_compareDICM}. The reason is that most of the images in LOL dataset are taken on indoor. And beyond that kinD++ shows striking illumination correction in the outdoor scene. However, in some cases like the last image in Fig. \ref{fig_compareLIME}, KinD++ gives unnatural enhancement results, and the denoising capacity is not attractive like DLN and our SurroundNet methods. We can see that the enhanced results reveal the validity of our model.

\section{Ablation experiments}
In order to demonstrate the effectiveness of each component of our model, we perform a series of comparative ablation studies.

\subsection{Transfer learning}
As mentioned above, our model is pre-trained on synthetic dataset. Several works\cite{sakaridis2018semantic}\cite{hahner2019semantic} have demonstrated that pre-training deep network in synthetic dataset and fine-tuning model in real dataset can achieve promising results. Such a technique is called transfer learning. We test our model in the cases of direct and transfer training. The results are shown in Table \ref{TABLE VI}. We can see that the transfer training version achieves 0.32 promotion on PSNR and 0.005 on SSIM.

\begin{table}[thbp]
\centering
\caption{comparison of the transfer and direct training strategies}
\label{TABLE VI}
\begin{tabular}{cccccc}
\hline
module &PSNR$\uparrow$ &SSIM$\uparrow$ &NIQE$\downarrow$ &LPIPS$\downarrow$ &Params$\downarrow$\\ \hline
Direct &21.77 & 0.835 &4.329 &0.224 &137k\\ \hline
Transfer &\textbf{22.09} &\textbf{0.840} &\textbf{4.311} &\textbf{0.221} &137k\\ \hline
\end{tabular}
\end{table}

\subsection{Effect of ASF}

ASF convolution kernel is the core component of our SurroundNet which helps to enhance low-light image in efficient manner. Here, we design experiments to prove the performance of our new convolution kernel on image enhancement. We first replace the ASF module with traditional $3\times3$ convolution layer. The comparison seems unfair because the ASF module uses the depth-wise convolution. It means ASF module has fewer parameters than $3\times3$ convolution. Therefore, we also conduct an experiment by using the $5\times5$ depth-wise convolution, which has a similar number of parameters with ASF convolution. Table \ref{table2} shows the experimental results. We can see that the ASF exceeds traditional convolution with 0.11 on PSNR and 0.07 on SSIM, meanwhile, ASF reduces about 21 percent of parameters. For the depth-wise convolution, ASF outperforms it with 0.23 on PSNR and 0.14 on SSIM. The result shows that our ASF kernel is high performance and lightweight, which is free lunch in low-light enhancement.

\begin{figure}[ht]
\centering
\includegraphics[width=1\linewidth]{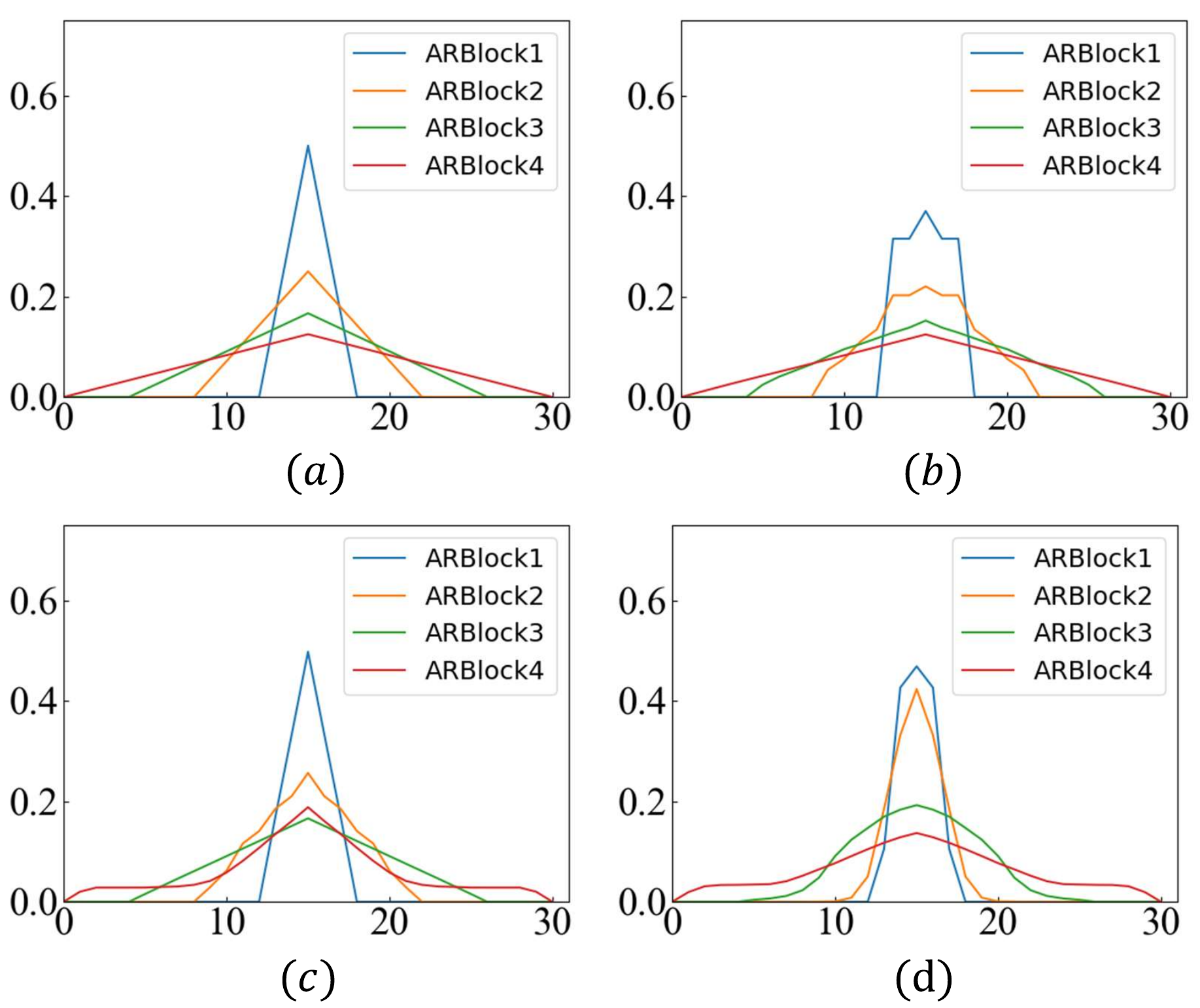}
\caption{The visualization of ASF kernel, (a) is the initial kernel weight, (b)(c)(d) are three random samples in different ARBlock.}
\label{fig_ASF_result}
\end{figure}

\begin{table}[htbp]
\centering
\caption{comparison of traditional, depthwise and our asf convolution kernel.}
\label{table2}
\begin{tabular}{cccccc}
\hline
module &PSNR$\uparrow$ &SSIM$\uparrow$ &NIQE$\downarrow$ &LPIPS$\downarrow$ &Params$\downarrow$\\ \hline
convolution &21.66 &0.828 &4.491 &0.240 &172k\\ \hline
DW convolution &21.54 &0.821 &4.571 &0.233 & 138k\\ \hline
ASF &\textbf{21.77} &\textbf{0.835} &\textbf{4.329} &\textbf{0.224} &\textbf{137k}\\ \hline
\end{tabular}
\end{table}
As we introduced in Section \ref{3b}, the ASF module can adaptively change the weight during the training process to get accurate illumination estimation. To verify this argument, we visualize the weight of the learned ASF kernel. As mentioned in Section \ref{3b}, we exploit two-dimension ASF by stacking two one-dimension kernels, so we only illustrate the one-dimension results as shown in Fig. \ref{fig_ASF_result}. We can see that the learned weight of ASFs appears more diverse and produces more flexible illumination estimation.

\subsection{Effect of ARBlocks}
In section \ref{arb}, we proposed the Adaptive Retinex Blocks (ARBlock) which extends the Single Scale Retinex to feature space. To verify the effectiveness of the ARBlocks, we firstly compare our SurroundNet with “Plain” net where we replace the ARBlocks with the stack of three $3\times3$ convolution blocks (i.e., conv.+ReLU). The “Plain” net structure, which is inspired by VGG\cite{simonyan2014very} is a popular network used in low-light enhancement tasks \cite{wei2018deep, zhang2021beyond}. As presented in Table \ref{table3}, SurroundNet outperforms PlainNet with 0.8 dB on PSNR and 0.012 on SSIM. The result indicates that ARBlock has the ability to improve performance in low light enhance task. To further analyze the ARBlock, we visualize the feature maps of the first ARBlock in the trained SurroundNet. And the results can be found in Fig. \ref{fig_ARB}. We can see that, due to the implementation of log function, the dynamic range of dark pixels will be stretched while bright part barely changes or gets compressed. It means the input will be lightened preliminary as shown in (Log-LL). Such explicit lighting operation simplifies the learning process. For ASF module in ARBlock, it can be regarded as an adaptive low-pass filter bank. The (I) part is the low-frequency part and (R) is the high-frequency part. We believe that such frequency decomposition benefits noise removal and image enhancement.

\begin{table}[htbp]
\centering
\caption{effect of arblocks}
\label{table3}
\begin{tabular}{cccccc}
\hline
module &PSNR$\uparrow$ &SSIM$\uparrow$ &NIQE$\downarrow$ &LPIPS$\downarrow$ &Params$\downarrow$\\ \hline
plainNet &20.97 &0.823 &4.789 &0.244 &159k\\ \hline
SurroundNet &\textbf{21.77} &\textbf{0.835} &\textbf{4.329} &\textbf{0.224} &\textbf{137k}\\ \hline
\end{tabular}
\end{table}

\subsection{The number of ARBlocks}
This work focuses on effective low-light image enhancement. As mentioned in Section \ref{arb}, most of the model parameters of our model belong to ARBlocks. Commonly, more blocks lead to better performance but are not always cost-effective. To explore the best trade-off between model size and enhance result, we train our model with different numbers of ARBlocks. The result can be seen from Table \ref{TABLE IV}. Both PSNR and SSIM indices climb, accompanied by the growth of the number of ARBlocks. However, the magnitude of improvement keeps reducing. Four-ARBlocks only exceeds Three-ARBlocks with 0.03 on PSNR and 0.04 on SSIM. Moreover, the NIQE index is even worse than the Three-ARBlocks. Stacking too much ARBlocks seems unworthy. In our paper, we use four ARBlocks for all experiments apart from this ablation comparison.

\begin{table}[htbp]
\centering
\caption{comparison on different number of arblocks}
\label{TABLE IV}
\begin{tabular}{cccccc}
\hline
module &PSNR$\uparrow$ &SSIM$\uparrow$ &NIQE$\downarrow$ &LPIPS$\downarrow$ &Params$\downarrow$\\ \hline
Block-1 &21.55 &0.827 &4.400 &0.242 &\textbf{61k}\\ \hline
Block-2 &21.53 &0.829 &4.358 &0.233 &86k\\ \hline
Block-3 &21.74 &0.831 &\textbf{4.301} &0.229 &111k\\ \hline
Block-4 &\textbf{21.77} &\textbf{0.835} &4.329 &\textbf{0.224} &137k\\ \hline
\end{tabular}
\end{table}

\subsection{Effect of Low-Exposure Denoiser}
\begin{figure}[ht]
\centering
\includegraphics[scale=0.4]{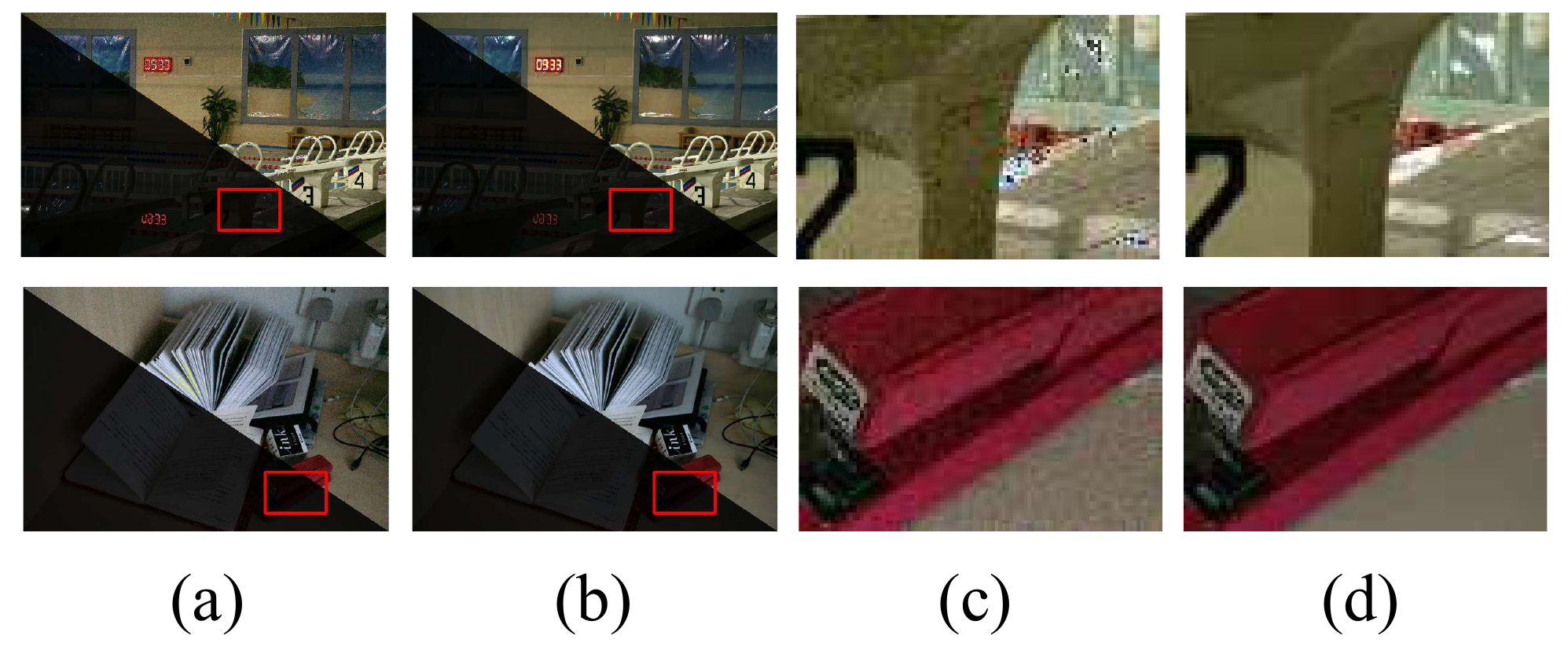}
\caption{Visualization on some output samples of Low-Exposure Denoiser. (a)(c) cols are the dark images, (b)(d) cols are LED module result, we light them by using 15$\times$ luminance gain(zoom for better view)}
\label{LEDshow}
\end{figure}

To reveal the validity of the LED module, we design two experiments: LES (Low-Exposure Supervision) and No-LES. LES means that the training loss contains the low-exposure supervision and No-LES does not. Both of them share the same network structure with the original SurroundNet. Moreover, we also test the performance in long training epochs case (3500 epoch). The result can be found in Table \ref{tablev}. We can see that, with 100 epochs, the difference between LES and No-LES is not obvious. The reason might be that a network always tries to learn illumination and color rather than noise in the early epochs. However, with 3500 epochs, LES surpasses the No-LES with 0.08 on SSIM, 0.003 on NIQE and 0.013 on LPIPS. The results confirm that LED module can smooth the image and help retain the image structure. We also visualize the output of LED module. Fig. \ref{LEDshow} exhibits some denoised samples. Fig. \ref{LEDshow}(b) shows LED module remains the color and illumination unchanged and the zoom-in result Fig. \ref{LEDshow}(d) displays LED can remove the noise and keep the edge sharp.
\begin{table}[htbp]
\centering
\caption{Comparison on low-exposure denoiser}
\label{tablev}
\begin{tabular}{cccccc}
\hline
module &PSNR$\uparrow$ &SSIM$\uparrow$ &NIQE$\downarrow$ &LPIPS$\downarrow$ &Params$\downarrow$\\ \hline
No-LES(100) &\textbf{21.83} &0.834 &\textbf{4.257} &0.227 &137k\\ \hline
LES(100) &21.77 &\textbf{0.835} &4.329 &\textbf{0.224} &137k \\ \hline \hline
No-LES/(3500) &\textbf{23.12} &0.842 &4.394 &0.209 &137\\ \hline
LES(3500) &22.71 &\textbf{0.850} &\textbf{4.391} &\textbf{0.196} &137k\\ \hline
\end{tabular}
\end{table}

\subsection{Effect of ECA}
To verify the function of ECA module, two SurroundNets are designed. We design a second SurroundNet without any channel feature aggregation module. The Table \ref{TABLE VII} shows that network improves performance with ECA. It is about 2.58 promotion on PSNR, 0.022 on SSIM, 0.112 on NIQE, and 0.022 on LPIPS. We can also see that ECA only increases model complexity slightly.

\begin{table}[htp]
\centering
\caption{effect of eca}
\label{TABLE VII}
\begin{tabular}{cccccc}
\hline
module &PSNR$\uparrow$ &SSIM$\uparrow$ &NIQE$\downarrow$ &LPIPS$\downarrow$ &Params$\downarrow$\\ \hline
w/o &19.19 &0.813 &\textbf{4.217} &0.246 &\textbf{134k}\\ \hline
w &\textbf{21.77} &\textbf{0.835} &4.329 &\textbf{0.224} &137k\\ \hline
\end{tabular}
\end{table}

\section{Conclusion and discussion}

In this work, we present a lightweight  SurroundNet for low-light image enhancement. It only takes less than 150k trainable parameters but achieves very competitive or even better results than SOTA methods. Different from previous Retinex-CNN based methods which employ stacks of convolution layers to estimate illumination, we propose ASF (Adaptive Surround Function) module to reach the same goal. ASF extends the concept of Surround Function in Retinex theory and makes the shape of Surround Function learned in a data-driven way. Furthermore, we propose Adaptive Retinex Block (ARBlock) which applies ASF on the feature space rather than RGB space. We also propose the LED method to smooth the image before enhancement. We evaluate the proposed method on the real-world low-light dataset. Experimental results demonstrate that the superiority of our submitted SurroundNet in both performance and network parameters against State-of-the-Art low-light image enhancement methods. 

In further work, we will focus on the global illumination information. We believe it can bring more generalization ability than the local one. In particular in the outdoor environment, which always has changeable lighting conditions. The interaction between noise and illumination is another interesting research content, it may be conducive to better trade-off between performance and parameters.

\bibliographystyle{plain}

\bibliography{ref}



\end{document}